\pdfoutput=1

\documentclass[11pt]{article}

\usepackage[]{emnlp2021}

\usepackage{times}
\usepackage{latexsym}

\usepackage[T1]{fontenc}

\usepackage[utf8]{inputenc}

\usepackage{microtype}

\title{Inaccessible Neural Language Models Could \\ Reinvigorate Linguistic Nativism}

\author{Patrick Perrine  \\
  California Polytechnic State University \\
  1 Grand Ave, San Luis Obispo, CA, 93410 \\
  \texttt{paperrin@calpoly.edu}}
  
\begin{document}
\maketitle


\section*{Abstract}
Large Language Models (LLMs) have been making big waves in the machine learning community within the past few years. The impressive scalability of LLMs due to the advent of deep learning can be seen as a continuation of empiricist lingusitic methods, as opposed to rule-based linguistic methods that are grounded in a nativist perspective. Current LLMs are generally inaccessible to resource-constrained researchers, due to a variety of factors including closed source code. This work argues that this lack of accessibility could instill a nativist bias in researchers new to computational linguistics, given that new researchers may only have rule-based, nativist approaches to study to produce new work. Also, given that there are numerous critics of deep learning claiming that LLMs and related methods may soon lose their relevancy, we speculate that such an event could trigger a new wave of nativism in the language processing community. To prevent such a dramatic shift and placing favor in hybrid methods of rules and deep learning, we call upon researchers to open source their LLM code wherever possible to allow both empircist and hybrid approaches to remain accessible.

\section{Introduction}

Large Language Models (LLMs) have been a popular topic of research among the academic community \cite{srivastava2022beyond}. The promise of a near-general purpose neural model for a variety of language processing tasks is indeed an attractive one \cite{xu2022systematic}. Deep learning has made significant developments in language tasks such as conversational language understanding \cite{tur2018deep}, spoken/text-based dialog systems \cite{celikyilmaz2018deep}, and natural language generation from images \cite{he2018deep}. Large Language models can be viewed as the natural progression away from the rigid rule-based systems that we've had since the 1950's \cite{chiticariu2013rule}, continuing the empiricist mentality of statistical natural language processing without the potentially costly and context-specific activity of feature engineering \cite{collobert2011natural}. However, with large corporations touting their ever-growing, state-of-the-art models under closed-source code and payment walls, it could be seen that these large language models are becoming \emph{less accessible}. Some organizations have acknowledged the potential harms that deep learning models could cause by establishing ethical frameworks \cite{ashurst2022ai} \cite{weidinger2022taxonomy}, but there are still growing concerns regarding accessibility and the result of false/irreproducible science \cite{kapoor2022leakage}. 

This criticism for empiricist methods is not new in linguistics-based science, in that Chomsky's Poverty of the Stimulus Argument \cite{stich1978empiricism} has a rich history of discussion and debate amongst linguists, scientists, and philosophers \cite{laurence2001poverty}. In this work, we will briefly introduce this debate over language learning between nativists and empiricists, relate these topics to research in natural language processing, and disucss how the current state of this research is reinforcing an imbalance between the two perspectives. We intend to deliver a neutral ground of analysis, as we agree that a hybrid approach for NLP research can lead to strong results. The current bias towards the highly-popular, but inaccessible empiricist methods utilizing LLMs could lead to a new wave of nativism in natural language processing work, following a large backlash against such empirical methods.

\section{Background}
We now provide a holistic background on the linguistic and scientific developments that encompass this issue.

\subsection{The Three Waves of Modern NLP}

We will give a brief background on the three main waves of modern natural language processing research: the rule-based theories popularized by Noam Chomsky \cite{chomsky1965aspects}, the statistics-based empiricist experiments \cite{jelinek1976continuous}, and today's popular methodology of deep learning for natural language processing \cite{collobert2011natural}. The first wave is considered to be under a nativist perspective \cite{laurence2001poverty}, whereas the latter waves are in support of an empiricist lens \cite{frank2019neural}.

\subsubsection{Rule-based NLP}

The concept of viewing language as a static system of rules to determine interpretation has been present as early as the 1830's \cite{humboldt1836uber}. Noam Chomsky popularized this perspective in the domain of linguistics as a challenge to an existing overbearance of empiricist methods \cite{chomsky1956three, laurence2001poverty}. 

This rule-based approach to linguistics dominated the field for decades, following Chomsky's mutliple works emphasizing and reinforcing this doctrine \cite{Chomsky1956logical, chomsky1957syntactic, chomsky1963formal, chomsky1965aspects, chomsky1968sound}. Being based in propositional logic and a fixed content, rule-based methods are arguably rather accessible to researchers with limited resources. These methods continued to be prevalent in the field until the 1970's, when statistical methods were proven to be very useful.

\subsubsection{Statistical NLP}

The roots of statistical language processing stem from Andrey Markov's efforts in computing bigram and trigram probabilities \cite{jurafsky2022speech} of vowel/consonant predictions using a novel as a corpus in 1913 \cite{markov2006example}. This $n$-gram approach was later applied to predicting sequences of English words \cite{shannon1948mathematical}. This popularized the notion of using Markov chains for use in a variety of applications within and outside of linguistics. 

Chomsky specifically challenged this use of finite-state Markov processes, the processes that formed $n$-gram based approaches, to be useless in serving as a comprehensive cognitive model of grammatical knowledge in humans \cite{chomsky1956three, chomsky1957syntactic, miller1963finitary}. This hindered the progress of probabilistic approaches in linguistics.

Over a decade later, statistical language processing was revitalized due in part to a series of successful experiments using $n$-gram models for speech recognition \cite{baker1975dragon, baker1975auto, jelinek1976continuous, bahl1983maximum, jelinek1990self}. These empiricist-based experiments showed that Chomsky's nativist theories do not extend to recognizing speech in real time as previously proposed \cite{chomsky1968sound}.

This marked a shift towards looking at language processing through an empirical lens, where a hypothesis test primarily guides the experimentation process, rather than theoretical insights \cite{manning1999foundations}. After the successful statistical speech recognition experiments of the mid 1970's, statistical NLP reigned as the dominant approach for decades. 

\subsubsection{ML-based NLP}
Researchers soon began to use shallow neural networks to reinforce statistical methodologies in NLP. In the late 2000's, the advent of deeper neural networks for NLP began to stir when scalable, hierarchical language models \cite{morin2005hierarchical, mnih2008scalable} and increased computing power became available for use by researchers. 

Alongside these developments, researchers became tiresome of having to hand-engineer features for neural networks to learn from, as this can be a costly and rather context-specific task \cite{collobert2011natural}. In was in the 2010's that deep learning became known more globally \cite{lecun2015deep}, with NLP being a highly prominent application for deep neural networks. This sparked the current practice of training large language models in efforts to create a general model for many language tasks \cite{srivastava2022beyond}. In essence, the empiricist era of NLP has persisted to today through the evolution of deep learning practices. Some applications of deep learning outside of language have even used empiricist terms such as \emph{tabula rasa} very openly \cite{silver2017mastering}. The use of deep neural networks for language tasks has been confirmed to reinforce empircist ideology \cite{frank2019neural}.

\section{Deep Learning Can Be Inaccessible}

Deep learning as a science has been under fire for a number of reasons. While there have been encouraging results across many application domains of deep learning and positive insights about their role in advancing empiricism \cite{buckner2018empiricism}, deep learning has garnered skepticsm from both in and outside of its community \cite{marcus2018deep, buckner2019deep}. 

These critcisms of deep NLP can stem from a \emph{lack of open sourcing of model code} and also data \cite{klein2017opennmt, fadel2019arabic, chen2021evaluating, guo2022freetransfer, xu2022systematic}. These issues are not exclusive to language processing, as other domains have reasons to leave aspects of their experimentation private or inaccessible when publishing \cite{siegle2015neural, suresha2018automated, farooq2020covid, zuin2020automatic, guo2022freetransfer}.

We now focus on issues with closed-source large language models due to their popularity and the recent claims of greater intelligence (even sentience), as opposed to other models \cite{y2022large}.

\section{Potential Harms}

\subsection{Potential Harms of Open-Sourcing LLMs}

To offer a well-rounded argument in favor of open-sourcing LLMs, we will briefly cover some intuitions behind close-sourcing them in terms of potential harms. 

LLMs could be repurposed for malicious purposes, particularly in generative tasks. LLMs have been seen to learn negative human biases/patterns in speech such as hate speech, discrimination, and the promotion of misinformation \cite{schramowski2022large}. If a powerful, pre-trained LLM is made open source, then it could be repurposed as an engine to cause harm across the internet at great scale \cite{weidinger2022taxonomy}. It could also be argued that open sourcing LLM code that has been deployed to end-users could pose security risks \cite{chang2020restricted}.

We counter the argument of potential LLM misuse by malicious parties by arguing that such models or derivatives of such should not be published in any form, open or closed source. We argue that LLM experimental papers that indicate such potential to cause harm at scale should be filtered out at the publication review stage, something that has been discussed in the deep learning community as of late \cite{ashurst2022ai}. We also counter the security concern argument by saying that this could hold true for all open source software that is deployable, not just LLMs.

\subsection{Potential Harms from Continued Close-Sourcing of LLMs}

We argue that there are more potential harms in the continued prevalence of close sourced LLM code than the potential harms of open sourcing them.

\subsubsection{Nativist Biases}

Given that LLM experiments are becoming so large, costly, and complex, it is difficult to argue that an independent researcher can stake a claim in this subfield. With top publication venues focusing heavily on empiricist experimentation \cite{russell2021artificial}, researchers outside the typical corporate scope of research could be incentivized to explore nativist, rule-based approaches to solve problems in the NLP domain. If it is the empiricist group's better interest to foster growth in their methodologies and not opposing methods, steps should be taken in order to make their approaches accessible. Also, for hybrid methods to function, an ML-based solution should be made accessible to combine with the ruleset from the nativist side. This trend could be fostering a new generation of Chomsky-following nativist NLP researchers, which would not bode well for empiricists if the public begins to lose interest in deep learning methods for NLP.

\subsubsection{Lack of Reproducibility}

We mention reproducibility and will further clarify its meaning due to an also recent, yet broader problem in deep learning research, the reproducibility crisis \cite{kapoor2022leakage}. Not only are large language models becoming difficult to reproduce, results from other areas of ML are becoming difficult to produce \cite{de2022eight, towers2022long}. Initiatives to measure reproducibility across publication venues have been created, such as the \href{https://www.cs.mcgill.ca/~jpineau/ICLR2018-ReproducibilityChallenge.html}{ML Reproducibility Challenge}. LLM experiments have been specifically reviewed to have a questionable about of reproducibility \cite{crane2018questionable, wieling2018reproducibility, cahyawijaya2022nusacrowd, silva2022no}. There is also implied to be a significant amount of computational irreproducibility of LLM experimentation, given model complexity and data, however, we leave this exploration for future work.

There is some hope in the form of positive reproducibility reports in deep learning \cite{gibson2022productive}. However, this growing amount of ``bad press'' for deep learning, specifically LLMs, could cause the public to begin distrusting LLM research. This, again, could trigger a revisiting of Chomsky's rule-based theories of language.

\subsubsection{Issues in NLP Education}

Given the previously mentioned issues, this lack of accessibility could affect the education of NLP methods. If students do not have access to code of LLMs, it could be difficult for them to learn to implement complex language model code of their own and learn to keep up with the state of the art. A lack of reproducibility could also be disenfranchising to a young, empircist NLP researcher, leading them to pursue nativist approaches. These issues could reinforce the use of statistical, pre-deep learning techniques in the classroom, but it is difficult to argue that publication venues are interested in shallow neural network experimentation at this time.

These issues combine to form an uneven playing field for students to study NLP in empiricist and hybrid forms. After studying NLP formally, they may be inclined to commit to nativist methods or even reinforce the popularity of them at scale.

\section{Potential Solution}

We ask that publication venues merit open source LLM experiments significantly higher than they do currently. We believe that this would mitigate the issues discussed previously in this work. There seem to be developments occuring now in the deep learning publication space to help implement this in a proper form of governance \cite{ashurst2022ai}.

\section{Conclusion}
In this work, we provided a comprehensive history of natural language processing methodologies over roughly the past century. We then used this narrative to lead into today's deep learning practices used in language processing, and current issues in an excessive closed sourcing of code for LLMs. It is our hope that this work inspires researchers and reviewers to champion open source language model code in order to pave the way for a more balanced research space.

\bibliographystyle{acl_natbib}
\bibliography{custom}

\end{document}